\documentclass[11pt]{article}

\usepackage{acl}

\usepackage{times}
\usepackage{latexsym}

\usepackage[T1]{fontenc}

\usepackage[utf8]{inputenc}

\usepackage{microtype}

\usepackage{inconsolata}

\usepackage{graphicx}

\usepackage{linguex} 

\usepackage{xcolor}
\usepackage{tcolorbox}
\usepackage{amsmath}

\usepackage{tabularx}
\usepackage{booktabs}

\usepackage{pgfplots}
\pgfplotsset{compat=1.17}

\definecolor{cQwen}{RGB}{31,119,180}      
\definecolor{cGemma}{RGB}{255,127,14}     
\definecolor{cMinistral}{RGB}{44,160,44}  
\definecolor{cGPT}{RGB}{148,103,189}      
  
\title{NER-D: A cross-domain benchmark for German NER}

\author{
  \textbf{Marc Brinner\textsuperscript{1}},
  \textbf{Philipp Cimiano\textsuperscript{2}},
  \textbf{Karolina Hanna Czok\textsuperscript{2}},
\\
  \textbf{Regina Stodden\textsuperscript{2}},
  \textbf{Viju Sudhi\textsuperscript{2}},
  \textbf{Sina Zarrie{\ss}\textsuperscript{1}}
\\
\\
  \textsuperscript{1}Computational Linguistics, Department of Linguistics, Bielefeld University
\\
   \textsuperscript{2}Center for Cognitive Interaction Technology, Faculty of Technology, Bielefeld University
\\
  {\small \texttt{\{marc.brinner,karolina.czok,regina.stodden,viju.sudhi,sina.zarriess\}@uni-bielefeld.de}}
\\
  {\small \texttt{cimiano@techfak.uni-bielefeld.de}}
}

\begin{document}
\maketitle
\begin{abstract}
We present NER-D, a cross-domain benchmark for German NER comprising 13 datasets spanning 9 domains, including General/Politics, Wikipedia, Legal, Medical, Historical, Financial, Agriculture and Archaeology. We examine the performance of both proprietary LLMs and state-of-the-art small open-source models using normal prompting and further examine the performance gains achievable via in-context learning and LoRA fine-tuning. Our results show that without any sort of optimization a proprietary model (GPT-5.4 mini) outperforms the smaller open-source models on 6 out of 13 datasets, but LoRA fine-tuning enables the smaller models to substantially outperform the proprietary baseline. We further analyze the influence of the domain and the tagset used in each task on the performance achieved in the different evaluation settings. Our benchmark will be available as a resource for the development of future German NER methods on both general and specialized domains.\footnote{We make our code, data, and models publicly available upon acceptance of this paper. }
\end{abstract}

\section{Introduction}
Large language models (LMs) have demonstrated strong performance across a variety of natural language processing tasks, including named entity recognition (NER) \cite{brown2020}. Proprietary models, such as those provided through commercial APIs, often achieve superior performance compared to open-weight alternatives. However, their practical adoption is limited by concerns regarding data privacy, lack of transparency and interpretability, dependence on third-party providers, and limited possibilities for model optimization \cite{Liao2024AI, Zhang2024, YAN2025100300}. In contrast, small open-weight LMs can be deployed on private infrastructure, allowing sensitive data to remain local. They can be customized through fine-tuning and incur no costs beyond the required infrastructure, making them an attractive alternative for privacy-sensitive applications and organizations requiring customizable, cost-effective NLP solutions \cite{Kukreja2024, manchanda2025opensourceadvantagelarge}.

NER is a well-researched task in natural language processing. It can be formulated as an information extraction or sequence labeling task in which text spans are annotated with predefined entity labels (e.g., person, location, organization) \cite{li2022}. While NER has been extensively studied for high-resource settings and general domains, research in specialized domains and languages other than English remains limited \cite{Krasadakis2024}. One major challenge in specialized domains requiring expert knowledge is the lack of annotated training data, often caused by the high effort required to design annotation guidelines and manually create high-quality annotations, as well as the high costs associated with expert annotation \cite{dao2025}. 
Examples \ref{example-ner-general} and \ref{example-ner-special} show the differences in  complexity of annotation between general-domain and specialized-domain NER.
\ex. \label{example-ner-general} 
\textcolor{blue}{Sundar Pichai}$_{\text{PER}}$ was born in 
\textcolor{purple}{India}$_{\text{LOC}}$ and is a CEO of
\textcolor{orange}{Google}$_{\text{ORG}}$.

\ex.  \label{example-ner-special} 
The patient with \textcolor{blue}{type 2 diabetes}$_{\text{DISEASE}}$
was treated with \textcolor{purple}{Metformin}$_{\text{DRUG}}$
but developed \textcolor{orange}{gastrointestinal intolerance}$_{\text{ADVERSE\_EFFECT}}$
with \textcolor{olive}{persistent nausea}$_{\text{SYMPTOM}}$.

In addition, privacy constraints in domains such as medicine can limit the availability and sharing of annotated datasets. For clinical German NLP, for example, \citet{FREI2022100212} have highlighted that only very scarce data is available due to privacy and other constraints. 

Despite the growing availability of German NER datasets from both general and specialized domains, these datasets are typically evaluated independently due to differences in annotation schemes and domain characteristics -- except \citet{pfister-hotho-2024-supergleber} which provide a German NLP benchmark across several tasks. Therefore, it remains unclear how well current LMs perform on German NER across specialized domains and in different settings (i.e., zero-shot, few-shot, and fine-tuning). 
To address this question, we propose NER-D 
as a new German NER benchmark that comprises 13 datasets from 9 different domains. We examine the performance of small open-source models compared to proprietary models (GPT-5.4 mini in particular) across all the datasets, investigating how big the gap is between proprietary and open-source models under different training regimens (zero-shot, few-shot and low-rank fine-tuning).

In particular, we address the following research questions:\vspace{-5px}
\begin{enumerate}
    \item[RQ1] How does NER performance vary across different domains?\vspace{-5px}
    \item[RQ2] How does the performance of small, open-weight LMs compare to proprietary SLM?\vspace{-5px}
    \item[RQ3] To what extent can few-shot learning and parameter-efficient fine-tuning reduce the performance gap between proprietary SLM and small, open-weight LMs?
\end{enumerate}

\section{Related Work} 

Several recent studies have investigated the capabilities of LMs for NER across different languages and domains. The Universal NER (UNER) benchmark \cite{blevins-etal-2026-universal} combines NER datasets from multiple languages using consistent annotation guidelines and evaluates both encoder-based models and LMs. 

For German, \citet{pfister-hotho-2024-supergleber} introduced SuperGLEBer, a broad benchmark for evaluating German LMs across multiple tasks, including sequence tagging tasks such as NER. For NER, five datasets in German were included, covering different domains: news (XGLUE; \citealp{liang-etal-2020-xglue}), politics (EUROPARL; \citealp{faruqui-pado-2010-training}), biomedical (BIOfid; \citealp{ahmed-etal-2019-biofid}), Wikipedia/news (\citealp{benikova-etal-2014-nosta}), and legal texts (\citealp{leitner-etal-2019-fine-grained}). However, this benchmark focuses only on the evaluation of German language models rather than comparing state-of-the-art proprietary and open-weight LMs. 

Other studies have investigated LMs for specialized domains. For example, in the comparative study of LMs for legal NER by \citet{deusser2024}, the authors evaluated eleven models on legal NER tasks across seven datasets in five languages, including English, Portuguese, German, Turkish, and Ukrainian. \citet{spiegel-etal-2025-adaption} investigated the use of 7-billion-parameter LMs for clinical NER in the German medical domain, evaluating their performance on three publicly available German gold-standard datasets. The authors additionally investigated instruction tuning with QLoRA and showed that it improves performance and reduces hallucinations. For historical texts, \citet{hiltmann2025ner4allcontextneedusing} evaluated GPT-4o on a manually curated historical dataset, testing zero-shot and few-shot approaches under different prompting strategies.

However, no previous work comprehensively evaluates German NER across multiple domains using both proprietary and open-weight models while comparing zero-shot, few-shot and fine-tuning approaches. This work addresses this gap by providing such an evaluation.


\section{Methods}
\subsection{Data}

We collected 13 publicly available NER datasets covering a variety of sources, including general text, legal documents, medical texts, historical documents, and specialized domains such as finance, agriculture, and archeology. The selection is not intended to be exhaustive but represents a diverse set of datasets that were identified through a review of available German NER resources. We selected datasets based on their availability, annotation quality, domain diversity, and suitability for evaluating language models. For multilingual datasets, we included only the German subset. 
The datasets included in our benchmark are summarized in \autoref{table-datasets} and briefly described below:

\paragraph{Europarl-NER} \cite{agerri-etal-2018-building,koehn-2005-europarl}
contains political proceedings from the European Parliament and provides general-domain entity annotations.\vspace{-5px}

\paragraph{SwissNER} \cite{swissner}
is a curated collection of 172 short, spoken-style German sentences, focusing on Switzerland.\footnote{As the dataset is very small with only 173 samples, we use it only for the experiments with zero-shot setting.}\vspace{-5px}

\paragraph{NoSta-D} \cite{benikova-etal-2014-nosta}
is a German NER dataset based on Wikipedia texts and annotated with general entity types as well as derivational and part-of entity categories.\vspace{-5px}

\paragraph{WikiBiography} \cite{filippova-strube-2006-using} is also based on Wikipedia texts, but only on bibliography pages. Compared to NoSta-D it contains fewer entity types which are more general.\vspace{-5px}

\paragraph{German LER} \cite{leitner-etal-2020-dataset}
is a legal NER dataset containing German legal documents from federal court decisions annotated with fine-grained legal entities.\vspace{-5px}

\paragraph{GerPS} \cite{feddoul-etal-2024-gerps}
contains German public service documents annotated with domain-specific entities related to administrative processes.\vspace{-5px}

\paragraph{GERNERMED} \cite{frei-etal-2023-german}
is a German medical NER dataset focusing on clinical entities related to medication.\vspace{-5px}

\paragraph{HUMADEX German NER} \cite{app15105585}
is a German medical NER dataset containing annotations for clinical entities such as problems, treatments, and tests.\vspace{-5px}

\paragraph{NER Staatsarchiv} \cite{staatsarchiv-ner}
contains historical administrative documents from Swiss archives annotated with general entity types.\vspace{-5px}

\paragraph{EHRI-NER} \cite{dermentzi-scheithauer-2024-repurposing}
is a historical dataset focused on Holocaust-related documents, containing annotations for general entities along with holocaust specific entities such as dates, camps, and ghettos.\vspace{-5px}

\paragraph{CO-Fun} \cite{foroutan-etal-2024-co}
is a financial-domain dataset annotated with entities related to company outsourcing.\vspace{-5px}

\paragraph{FAIRagro} \cite{abdelmalak_fairagro_ner_2025}
is an agriculture-domain dataset created for metadata enrichment and contains fine-grained annotations related to agricultural research.\vspace{-5px}

\paragraph{Archaeo-NER} \cite{brandsen_archaeo_ner_2025}
is an archaeology-domain dataset containing annotations for archeological entities.\vspace{-5px}

\medskip

To enable a consistent comparison across datasets, all resources were converted into a unified format, which includes splitting larger samples into sentences that are individually processed by the LLMs. The processed datasets will be released in a HuggingFace collection. 

After processing, each sample consists of an input sentence and an output JSON object containing all possible entity types defined for the respective dataset as keys. The value of each key is a list of entity mentions found in the sentence; if no entity of a given type is present, the list remains empty. This format is identical across all datasets, while the set of available entity types depends on the original annotation scheme of each resource. In addition, duplicated entity mentions within individual labels were removed. We used the original dataset splits whenever they were provided by the authors. For datasets without predefined splits, we created training, validation, and test splits using an 80/10/10 ratio.

\begin{table*}[t]
\centering
\small
\begin{tabularx}{\textwidth}{lllXl}
\toprule
\textbf{Dataset} & \textbf{Domain} & \textbf{Train/Val/Test} & \textbf{Labels} \\
\midrule

Europarl-NER & General (politics) &
639 / 80 / 80 &
4: PER, ORG, LOC, MISC \\

SwissNER & General (Swiss) &
- / - / 172 &
13: Canton, City, Company, Country, Event, Lake, Landmark, Mountain, Organization, Person, Region, River, Village \\

NoSta-D & Wikipedia &
24000 / 2200 / 5100 &
12: PER, PERpart, PERderiv, LOC, LOCpart, LOCderiv, ORG, ORGpart, ORGderiv, OTH, OTHpart, OTHderiv \\

WikiBiography & Wikipedia &
19046 / 2337 /  2162 &
5: LOC, ORG, PERSON, TEMP, UNDEF\_NE \\
 
German LER & Legal &
22754 / 2732 / 2893 &
16: PER, RR, AN, LD, ST, STR, ORG, UN, INN, GRT, MRK, GS, VO, EUN, VS, VT \\

GerPS & Legal &
10718 / 595 / 596 &
10: Mitwirkender, Handlungsgrundlage, Ergebnisempfänger, Hauptakteur, Aktion, Dokument, Signalwort, Frist, Bedingung, Datenfeld \\

GERNERMED & Medical &
25579 / 1421 / 1422 &
7: Drug, Route, Strength, Frequency, Duration, Form, Dosage \\

HUMADEX German NER & Medical &
381225 / 21179 / 21180 &
3: PROBLEM, TREATMENT, TEST \\

NER Staatsarchiv & Historical &
3986 / 221 / 222 &
6: PER, LOC, ORG, PERderiv, LOCderiv, ORGderiv \\

EHRI-NER & Historical &
3951 / 219 / 220 &
6: PER, LOC, ORG, DATE, CAMP, GHETTO \\

CO-Fun & Financial &
758 / 94 / 96 &
4: Auslagerung, Ort, Software, Unternehmen \\

FAIRagro & Agriculture &
581 / 65 / 54 &
15: Crop species, Crop variety, Soil texture, Soil reference group, Soil depth, Bulk density, pH value, Organic carbon, Available nitrogen, Location name, Latitude, Longitude, Start time, End time, Duration \\

Archaeo-NER & Archaeology &
7704 / 3364 / 3161 &
7: Artefacts, Contexts, Locations, Materials, Time Periods, Species \\

\bottomrule
\end{tabularx}\vspace{-5px}
\caption{Overview of datasets included in the NER-D benchmark.}
\label{table-datasets}
\end{table*}\vspace{-5px}

\subsection{Models}

We investigate a total of four models spanning different accessibility levels, architectures, sizes, and training objectives.

The first two are the leading open-weight LMs in the smaller size class ($\leq$12B parameters): 
Qwen3.5-9B \cite{qwen3.5}\footnote{\url{https://huggingface.co/Qwen/Qwen3.5-9B}} and Gemma4 12B Instruct~\cite{gemma-4}\footnote{\url{https://huggingface.co/google/gemma-4-12B-it}}. We additionally evaluate Ministral 8B Instruct\footnote{\url{https://huggingface.co/mistralai/Ministral-8B-Instruct-2410}}, which we selected for its explicit focus on multilinguality, including German. 

We explicitly chose not to evaluate any models developed specifically for German, as the available options are already outdated relative to these state-of-the-art variants, so that the inferior general task-solving capabilities would constitute a severely limiting factor. Further, multilinguality (thus including the comprehension of German texts) has become standard across all frontier open-weight models, including those evaluated in this work.

Finally, as an additional baseline, we include OpenAI's GPT-5.4-mini~\cite{gpt-5.4} to assess whether cost-effective closed-source models offer any substantial advantage over their open counterparts.

\subsection{Experimental Setup}

For each dataset, we created a German task-specific prompt describing the available entity types and the required output format (see Appendix \ref{appendix:prompt}). All prompts followed the same template and differed only in the manually curated entity definitions corresponding to the specific dataset. 
Models were instructed to return a valid JSON object containing all entity types, with empty lists for entity types not present in the input sentence. We intentionally did not perform prompt optimization. Instead, we used a simple, standardized prompting strategy to evaluate model performance under realistic conditions, where extensive dataset-specific prompt development is often impractical and requires substantial manual effort and domain expertise.

All models were evaluated on the test split of each dataset. We report results in both zero-shot and few-shot settings. In the zero-shot setting, models received only the task instructions and the input sentence. In the few-shot setting, five demonstration examples sampled from the validation split were included in the prompt. The same prompt template, inference settings and evaluation procedure were used for all models to ensure a fair comparison.

\subsection{Fine-Tuning}

For the three open-weights models, we also evaluate the improvements in downstream performance that can be realized via supervised fine-tuning. Since all models possess basic capabilities for solving the different tasks already, we experiment with performing a low-rank adaptation \cite{hu2022lowrank} of the original model weights on all linear layers ($r=16$, $\alpha=32$, $\textrm{dropout}=0.05$, $\textrm{lr}=1e-4$, $\textrm{batch size} = 128$).

Models are trained with the same prompt format used at inference, and the loss is computed on the target tokens only, thus optimizing the model solely for producing the correct target JSON.

Model selection and early stopping rely on determining the loss on the held-out validation split. This validation is done after every epoch, or (for larger datasets) after every 10k training examples, and training is stopped if the validation loss did not decrease for three consecutive evaluations.








\subsection{Evaluation}

For each class, the model is tasked to predict a JSON object containing all class names as keys, and the corresponding list of detected entities for that class as values. To evaluate the prediction, we parse this JSON from the model's response, treating a JSON object that cannot be parsed as predicting no entities.

The list of predicted entities are then compared to the ground-truth annotations to calculate the number of entities a) that were both predicted and annotated (\textit{true positives}), b) entities that were predicted but are not present in the ground truth list (\textit{false positives}) and c) entities that belong to the ground truth but were not predicted by the model (\textit{false negatives}). 

These individual, sample-specific true positives, false positives, and false negatives are then accumulated over all entities of all types across the entire test set, and both micro and macro F1 scores are computed from these corpus-level counts.

To evaluate both the broad recognition, but also the exact, verbatim matching of each ground-truth entity, we report two variants of this F1 score that differ only in how a predicted entity is matched to a gold entity.

In the \textit{exact-match F1}, a predicted string counts as a true positive only if it is identical to a gold string of the same type. Because generative models frequently produce near-misses that correctly identified the entity but differ superficially (e.g. an included article or a slightly different span boundary), we additionally report a \textit{partial-match F1}: within each type, predicted and gold strings are matched one-to-one (greedily, highest similarity first) and a pair counts as a match if their token-set similarity (a token-level Levenshtein ratio) is at least 0.90. The partial-match score thus upper-bounds the exact-match score and quantifies how much of the gap is due to superficial string differences rather than genuine extraction errors.


\definecolor{cMinistral}{HTML}{4C72B0}
\definecolor{cQwen}{HTML}{DD8452}
\definecolor{cGemma}{HTML}{55A868}
\definecolor{cGPT}{HTML}{8172B3}

\pgfplotsset{
  base bar/.style={ybar, bar width=16pt, bar shift=#1},
  light/.style={draw=#1!55, fill=#1!28, forget plot},
  dark/.style={draw=#1!60!black, fill=#1, area legend},   
}

\begin{figure*}[t]
  \centering
  \begin{tikzpicture}
    \begin{axis}[
        width=\textwidth, height=5.5cm,
        ymin=0, ymax=0.9,
        ytick={0,0.2,0.4,0.6,0.8},
        yticklabel style={font=\small},
        ylabel={Average entity-level F1},
        y label style={font=\small},
        symbolic x coords={Zero-shot, Few-shot, Fine-tuning},
        xtick=data,
        x tick label style={font=\small},
        enlarge x limits=0.18,          
        axis x line*=bottom,
        axis y line*=left,
        axis line style={gray!60},
        tick style={gray!60},
        ymajorgrids=true,
        major grid style={line width=0.4pt, dotted, gray!45},
        clip=false,
        legend cell align=left,
        legend style={
          at={(0.5,1.04)}, anchor=south,
          legend columns=-1, draw=none, font=\small,
          /tikz/every even column/.append style={column sep=10pt},
        },
      ]

      \addplot[base bar=-30pt, light=cMinistral]
        coordinates {(Zero-shot,0.607) (Few-shot,0.626) (Fine-tuning,0.800)};
      \addplot[base bar=-30pt, dark=cMinistral]
        coordinates {(Zero-shot,0.430) (Few-shot,0.470) (Fine-tuning,0.732)};
      \addlegendentry{Ministral 8B}

      \addplot[base bar=-10pt, light=cQwen]
        coordinates {(Zero-shot,0.624) (Few-shot,0.663) (Fine-tuning,0.813)};
      \addplot[base bar=-10pt, dark=cQwen]
        coordinates {(Zero-shot,0.481) (Few-shot,0.542) (Fine-tuning,0.747)};
      \addlegendentry{Qwen 3.5 9B}

      \addplot[base bar=10pt, light=cGemma]
        coordinates {(Zero-shot,0.663) (Few-shot,0.688) (Fine-tuning,0.799)};
      \addplot[base bar=10pt, dark=cGemma]
        coordinates {(Zero-shot,0.507) (Few-shot,0.560) (Fine-tuning,0.731)};
      \addlegendentry{Gemma 4 12B}

      \addplot[base bar=30pt, light=cGPT]
        coordinates {(Zero-shot,0.658)};
      \addplot[base bar=30pt, dark=cGPT]
        coordinates {(Zero-shot,0.504)};
      \addlegendentry{GPT-5.4 mini}

    \end{axis}
  \end{tikzpicture}\vspace{-10px}
  \caption{Average entity-level F1 across all datasets per model, grouped by
           prompting/training strategy. The dark segment of each bar shows the
           exact-match micro-F1; the lighter extension shows the score under
           partial (fuzzy) matching. GPT-5.4 mini was evaluated zero-shot only.}
  \label{fig:avg-scores}
\end{figure*}
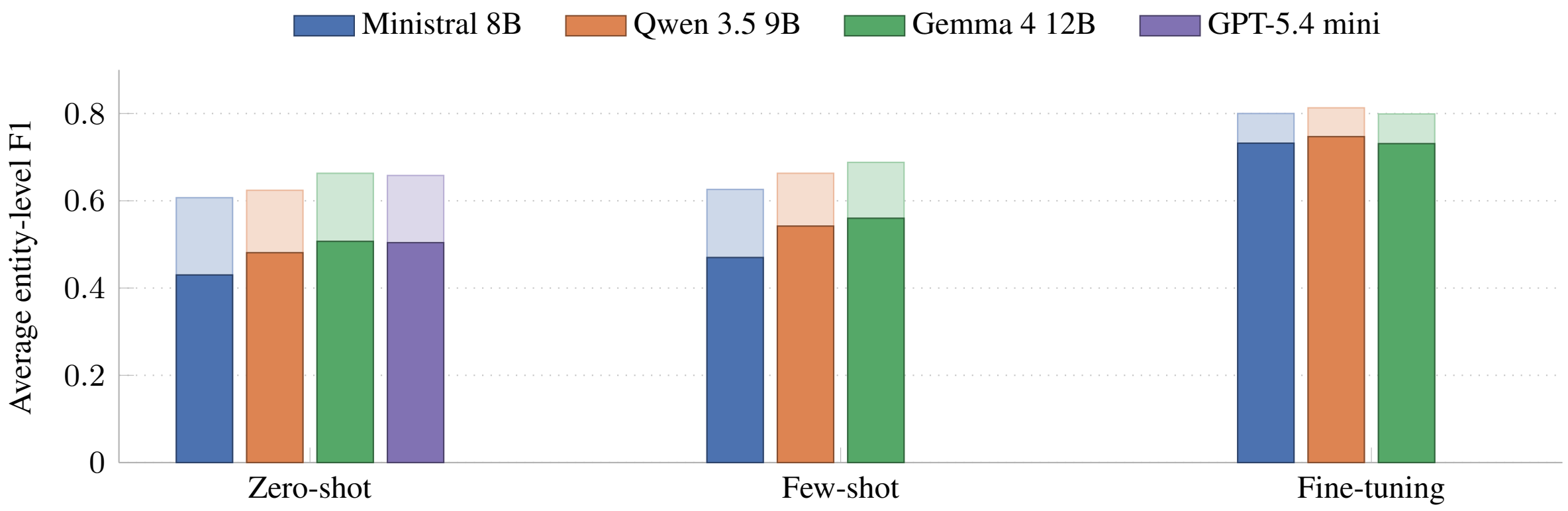

\begin{table*}[t]
  \centering
  \resizebox{\textwidth}{!}{%
  \begin{tabular}{l ccc ccc ccc c}
  \toprule
   & \multicolumn{3}{c}{\textbf{Ministral 8B}} & \multicolumn{3}{c}{\textbf{Qwen 3.5 9B}}
   & \multicolumn{3}{c}{\textbf{Gemma 4 12B}} & \textbf{GPT-5.4 mini} \\
  \cmidrule(lr){2-4} \cmidrule(lr){5-7} \cmidrule(lr){8-10} \cmidrule(lr){11-11}
  \textbf{Dataset} & 0-shot & 5-shot & LoRA & 0-shot & 5-shot & LoRA & 0-shot & 5-shot & LoRA & 0-shot \\
  \midrule 

FAIRagro & 0.492 & 0.523 & 0.731 & 0.582 & 0.692 & \textbf{0.857} & 0.484 & 0.542 & \underline{0.769} & 0.441 \\
Archaeo NER & 0.412 & 0.450 & \underline{0.679} & 0.471 & 0.487 & \textbf{0.681} & 0.516 & 0.534 & 0.671 & 0.476 \\
CO-Fun & 0.652 & 0.698 & \textbf{0.869} & 0.639 & 0.692 & \underline{0.844} & 0.619 & 0.711 & 0.829 & 0.614 \\
WikiBiography & 0.434 & 0.428 & \textbf{0.856} & 0.555 & 0.590 & \underline{0.840} & 0.539 & 0.618 & 0.837 & 0.584 \\
Europarl-ner & 0.328 & 0.387 & 0.688 & 0.460 & 0.540 & \underline{0.754} & 0.648 & 0.708 & \textbf{0.761} & 0.603 \\
SwissNER & 0.821 & -- & -- & \underline{0.849} & -- & -- & 0.788 & -- & -- & \textbf{0.862} \\
NoSta-D & 0.504 & 0.527 & 0.857 & 0.578 & 0.634 & \textbf{0.882} & 0.658 & 0.661 & \underline{0.875} & 0.637 \\
EHRI-NER & 0.549 & 0.591 & \textbf{0.845} & 0.603 & 0.625 & 0.835 & 0.629 & 0.656 & \underline{0.843} & 0.674 \\
NER Staatsarchiv & 0.583 & 0.646 & 0.853 & 0.684 & 0.725 & \textbf{0.888} & 0.718 & 0.761 & \underline{0.881} & 0.766 \\
German LER & 0.309 & 0.379 & \textbf{0.622} & 0.217 & \underline{0.427} & 0.401 & 0.276 & 0.400 & 0.405 & 0.239 \\
GerPS & 0.146 & 0.161 & 0.308 & 0.182 & 0.201 & \underline{0.353} & 0.181 & 0.206 & \textbf{0.369} & 0.168 \\
GERNERMED & 0.521 & 0.588 & 0.799 & 0.539 & 0.637 & \textbf{0.955} & 0.539 & 0.634 & \underline{0.887} & 0.561 \\
HUMADEX German NER & 0.229 & 0.258 & \textbf{0.684} & 0.257 & 0.258 & \underline{0.679} & 0.275 & 0.286 & 0.642 & 0.281 \\ \midrule
Average & 0.430 & 0.470 & \underline{0.732} & 0.481 & 0.542 & \textbf{0.747} & 0.507 & 0.560 & 0.731 & 0.504 \\

  \bottomrule
  \end{tabular}%
  }\vspace{-5px}
  \caption{Micro-averaged entity-level F1 on the German NER benchmark. Per dataset,
  the best score is \textbf{bold} and the second best \underline{underlined}.
  \emph{0-shot}: zero-shot, \emph{5-shot}: 5-shot in-context, \emph{LoRA}: fine-tuned.
  The final row reports per-column averages over the datasets, excluding SwissNER, which is only evaluated in the 0-shot setting due to the small dataset size that does not allow for creating train or validation splits.}
  \label{tab:ner-micro-f1}
  \end{table*}

\begin{table*}[t]
  \centering
  \resizebox{\textwidth}{!}{%
  \begin{tabular}{l ccc ccc ccc c}
  \toprule
   & \multicolumn{3}{c}{\textbf{Ministral 8B}} & \multicolumn{3}{c}{\textbf{Qwen 3.5 9B}}
   & \multicolumn{3}{c}{\textbf{Gemma 4 12B}} & \textbf{GPT-5.4 mini} \\
  \cmidrule(lr){2-4} \cmidrule(lr){5-7} \cmidrule(lr){8-10} \cmidrule(lr){11-11}
  \textbf{Dataset} & 0-shot & 5-shot & LoRA & 0-shot & 5-shot & LoRA & 0-shot & 5-shot & LoRA & 0-shot \\
  \midrule
FAIRagro & 0.677 & 0.677 & 0.808 & 0.727 & 0.731 & \textbf{0.857} & 0.710 & 0.712 & \underline{0.846} & 0.610 \\
Archaeo NER & 0.576 & 0.604 & \textbf{0.788} & 0.607 & 0.627 & \underline{0.784} & 0.649 & 0.669 & 0.760 & 0.617 \\
CO-Fun & 0.840 & 0.856 & \textbf{0.909} & 0.780 & 0.818 & \underline{0.889} & 0.793 & 0.866 & 0.870 & 0.768 \\
WikiBiography & 0.571 & 0.571 & \textbf{0.873} & 0.635 & 0.677 & \underline{0.857} & 0.600 & 0.680 & 0.855 & 0.646 \\
Europarl-ner & 0.463 & 0.495 & 0.766 & 0.633 & 0.657 & \underline{0.783} & 0.745 & 0.764 & \textbf{0.851} & 0.679 \\
SwissNER & 0.831 & -- & -- & \underline{0.856} & -- & -- & 0.812 & -- & -- & \textbf{0.867} \\
NoSta-D & 0.582 & 0.588 & 0.875 & 0.619 & 0.675 & \textbf{0.900} & 0.714 & 0.704 & \underline{0.896} & 0.678 \\
EHRI-NER & 0.734 & 0.788 & \textbf{0.902} & 0.743 & 0.757 & 0.891 & 0.773 & 0.797 & \underline{0.900} & 0.777 \\
NER Staatsarchiv & 0.788 & 0.807 & 0.890 & 0.833 & 0.835 & \textbf{0.928} & 0.797 & 0.856 & \underline{0.923} & 0.842 \\
German LER & 0.487 & 0.518 & \textbf{0.735} & 0.296 & 0.522 & \underline{0.688} & 0.483 & 0.463 & 0.556 & 0.572 \\
GerPS & 0.275 & 0.273 & 0.346 & 0.291 & 0.309 & \underline{0.403} & 0.319 & 0.339 & \textbf{0.411} & 0.311 \\
GERNERMED & 0.793 & 0.814 & 0.896 & 0.810 & 0.829 & \textbf{0.973} & 0.813 & 0.834 & \underline{0.938} & 0.834 \\
HUMADEX German NER & 0.501 & 0.526 & \textbf{0.807} & 0.513 & 0.513 & \underline{0.805} & 0.565 & 0.568 & 0.782 & 0.561 \\ \midrule
Average & 0.607 & 0.626 & \underline{0.800} & 0.624 & 0.663 & \textbf{0.813} & 0.663 & 0.688 & 0.799 & 0.658 \\
  \bottomrule
  \end{tabular}%
  }\vspace{-5px}
  \caption{Partial-match (fuzzy) entity-level F1 on the German NER benchmark, using
  token-set-ratio matching at threshold 0.90. Per dataset, best \textbf{bold},
  second best \underline{underlined}; final row = per-column averages.}
  \label{tab:ner-partial-f1}
  \end{table*}\vspace{-5px}

\section{Results \& Discussion}

The results for the exact-match micro F1 and partial-match micro F1 are listed in Tables \ref{tab:ner-micro-f1} and \ref{tab:ner-partial-f1}, respectively, and visualized in Figure \ref{fig:avg-scores}. Macro F1 scores are included as additional reference in Appendix~\ref{app:macro}.

\subsection{Domain-specific Analysis}
First, we analyze difference of performance across different domains. Interestingly, while partial-match micro F1 scores vary widely from 0.275 to 0.867 in the zero-shot setting, we do not observe a significantly better performance in more general domains like Wikipedia or political texts compared to more specialized domains like archaeology or agriculture, even though these specialized domains are likely covered less extensively during model training. Further, the specified entity labels and jargon of the datasets of these specialized domains do not seem to harm the quality of the models' prediction. 
However, for some specialized domain datasets like GERPS (and other legal datasets in general) the scores are significantly lower compared to the other datasets, which we analyze more extensively in Section~\ref{sec:error_analysis}.

Further, the evaluation identifies some variance across datasets within the same domain. For example, models perform much better on GERNERMED than on HUMADEX (both medical), although the latter has a lower number of different entity labels. However, fine-tuning on the large training dataset helps to reduce this performance gap between the two medical datasets to some extent, with a similar pattern arising in the legal domain. Within other domains such as historical texts the scores are quite similar across datasets, thus ultimately suggesting that the models' performance is more dependent on specific properties of the the dataset (e.g., its label set and annotation procedure) than on the domain (RQ1). 


\subsection{Small, open LMs vs. closed LMs}
\paragraph{Zero-Shot Performance}
Both the open models and the proprietary model show reasonable zero-shot performance, with average exact-match F1 scores being in the range of $0.430$-$0.507$. This demonstrates a basic capability of solving NER tasks even in German, even though substantial differences between the different tasks exist. For a more detailed discussion of task-specific difficulties, see Section \ref{sec:error_analysis}.

Comparing the open-models with the proprietary model (RQ2), GPT-5.4 mini interestingly does not outperform the - likely smaller - open weights models, with Gemma 4 12B surpassing GPT-5.4 mini in average zero-shot exact-match performance (0.507 vs. 0.504). However, no model is able to consistently outperform the others: while GPT-5.4 mini performs best on 6 out of 13 datasets, Qwen manages to outperform the other models on 3 datasets, with Gemma and Ministral leading on 2 datasets each.

Thus, while general trends in average performance over all datasets are evident, individual datasets display high variance between models so that a dataset-specific choice of model is mandatory for optimal zero-shot performance.

\paragraph{Few-Shot Performance}

Few-shot prompting provides additional task-specific information by including a small number of correct input-output examples within the model's context during generation. In this way, the model behavior can be adapted to the target task without requiring any modification of its parameters.

Indeed, including five demonstrations in the prompt leads to substantial improvements in exact match performance (e.g., Qwen: 0.481 $\rightarrow$ 0.542, Gemma: 0.507 $\rightarrow$ 0.560), thus showing that the models are able to draw inferences from the examples that lead to reliable performance improvement.

Interestingly, these improvements are much less pronounced for the partial-match F1 score, indicating that the model does not become substantially better at recognizing entities themselves. Instead, the larger gains in exact-match performance suggest that the demonstrations primarily help the model learn task-specific annotation conventions, such as whether articles or other surrounding words should be included in entity spans. Such conventions can often be learned from just a few examples, while precise understanding of classes is usually more difficult to acquire.

\paragraph{Effects of Fine-Tuning}

In contrast to the few-shot setting, fine-tuning allows for stronger adaptation of model behavior to a specific task. Further, it allows the model to implicitly take information from hundreds of samples into account, in contrast to the much more limited number in the few-shot setting.

This manifests in substantially improved performance, with exact match scores increasing on average by 0.264 across all three models. (Ministral: 0.430 $\rightarrow$ 0.732, Qwen: 0.481 $\rightarrow$ 0.747, Gemma: 0.507 $\rightarrow$ 0.731). In contrast, partial match scores increase substantially as well, but by a lower 0.173 on average (Ministral: 0.607 $\rightarrow$ 0.800, Qwen: 0.624 $\rightarrow$ 0.813, Gemma: 0.663 $\rightarrow$ 0.799).

Ultimately, fine-tuning therefore also leads to a narrower gap between exact-match and partial-match performance, thus indicating that the fine-tuned model is better at matching span-boundaries of ground-truth entities exactly and has thus learned the annotation conventions from the training data.

To answer research question 3, fine-tuning not only helps surpassing the simpler in-context learning method, but also makes small, local models substantially outperform the GPT-model by over 0.2 points in exact-match F1 if averaged across all datasets. Further, this performance gap is consistent across all datasets, so that even small numbers of annotated training samples are sufficient to make local models the most reliable choice.

Lastly, a notable insight is that average performance after fine-tuning is very similar for all three models. Thus, while in the untrained zero-shot setting choosing the right model was crucial, performance differences are often less pronounced after tuning parameters with supervised data.

\subsection{Error Analysis}
\label{sec:error_analysis}

\paragraph{Difficult Datasets}

Although all models demonstrate reasonable performance even in the zero-shot setting, thus indicating a baseline capability for solving German NER tasks, some datasets prove substantially more challenging than others.

The GerPS dataset, for instance, consistently yields comparatively poor results across both zero-shot and few-shot settings, with the best-performing models achieving a partial-match F1 score only slightly above 0.3. Moreover, fine-tuning only leads to small improvements, which suggests that there are structural factors that substantially limit downstream performance.

When analyzing failure modes on this dataset, we found that subpar performance can, at least in part, be attributed to the standardized preprocessing pipeline employed in our experiments, which involved splitting longer documents into individual sentences for both training and evaluation. However, the original GerPS task requires identifying semantic roles within German administrative texts, which often depend on information distributed across the full document. By removing this broader context, sentence-level processing imposes an inherent upper bound on achievable performance.

Further, some labels (e.g., "Bedingung"; EN: "condition") expect very long sequences, often spanning several clauses. Such labels prove to be more difficult, especially before fine-tuning. In comparison, high scoring datasets contain mostly annotations with single-word sequences (i.e. SwissNER) or only a small label set (i.e., CO-FUN). 

Another dataset with weaker performance, particularly prior to fine-tuning, is German LER. In contrast to GerPS, this behavior appears to be primarily caused by the inherent complexity of the task itself. The dataset focuses on the legal domain and contains a fine-grained entity schema that divides semantically similar entities into different classes. In practice, we find that the zero-shot models are not able to perform this precise distinction and tend to assign each entity to too many classes. While recall therefore is very high, the precision drops and the resulting F1 score is mediocre. Fine-tuning then improves performance substantially by helping the models distinguish better between relevant and irrelevant classes for each entity.

\paragraph{Prompt Limitations}

Particularly in the zero-shot and few-shot settings, the prompt itself is a major driver of model performance, and thus also has the potential to be an important limiting factor. Usually, prompts provide only a concise task description and therefore cannot capture every aspect of a dataset's annotation guidelines or class definitions, which can negatively affect downstream performance.

One effect resulting from an under-specification of annotation rules within the prompt is the consistently higher partial-match F1 scores compared to their exact-match counterparts. As discussed previously, models often succeed in identifying the correct entity but fail to identify the precise span boundaries of the ground truth annotations, since the prompt does not contain sufficient information to decide between multiple different valid annotation conventions. In some cases, relatively small prompt refinements could alleviate this issue, for example by explicitly specifying whether determiners should be included in entity spans (e.g., "the European Court" versus "European Court") or whether honorifics and name prefixes should be annotated (e.g., "Mr. Smith" versus "Smith").

However, precisely specifying annotation rules for all edge cases is usually infeasible. As a result, despite the increasingly strong zero-shot capabilities of modern LLMs, supervised fine-tuning remains the most reliable approach for achieving optimal performance.

A second limitation concerns the entity class definitions themselves. Some datasets contain intentionally broad or heterogeneous categories that cannot be adequately described within a single sentence. For example, the NoSta-D dataset includes an \emph{Other} category that encompasses multiple types of entities not covered by the other labels. As a result, zero-shot models struggle to apply it consistently based on the under-specified class description. Again, supervised fine-tuning enables the model to learn these implicit distinctions directly from annotated examples, resulting in substantially improved performance.



\section{Conclusion \& Future Work}
In this benchmark called NER-D, we provided an overview of how well small, open language models perform on the task of German named entity recognition. NER-D is the first benchmark that focuses on German NER across various domains, spanning from a general domain like Wikipedia to specialized domains like agriculture or archaeology. Overall, our results show that multilingual, small, open LMs already in a zero-shot setting have at least basic capability of solving NER tasks even in German.

Our results show that open SLMs perform differently across datasets, but the performance differences cannot be explained based on the domain specialization. For low-resource domains such as agriculture or archaeology, the performance of the models is comparable to the performance on other more general domains. The complexity of the NER tasks seems to be more dependent on the annotation procedure than on the jargon on specialized labels of the domains.

We also established that small, open LMs can compete with proprietary SLMs when enough samples are available for fine-tuning. Although creating training data on specialized domains is very time consuming, we have shown that it is worth to let the model learn the task- and domain specific annotation conventions. Thus, small, open models show benefits compared to closed SLMs not only with respect to data privacy and lower resource costs but also with respect to perfomance improvements. 

In future work, we plan to compare different sizes of models within the same model family to see whether our results also hold across very small models (e.g., smaller than 3B). Further, we have focused more on generalizability across several domains than on the best performance on each dataset or domain. In future work, we believe that the performance of the open SLMs can even be increased by optimizing the prompt per domain or set of entity labels.


\section*{Limitations}
In this section, we discuss a few limitations of our work. First, we only considered multilingual language models. It would be interesting to compare our results also to newer German-only LMs. 

Further, including always 5 randomly-picked samples in the few-shot setting or providing less details of annotation guideline in the prompt could have limited the performance of the LMs. 

Lastly, we  acknowledge a limitation of the number of specialized domains compared to more general domains in the benchmark, which is mostly due to the lack of publicly available German NER datasets on specialized domains. 


\bibliography{references/custom}

\appendix

\section{Appendix}
\label{sec:appendix}
\subsection{Prompt Template}
\label{appendix:prompt}
We used the following prompt template in German for all datasets.

\begin{quote}

  Du bist ein präzises System zur Named Entity Recognition (NER) für deutsche Texte. Deine Aufgabe ist es, strukturierte Entitäten basierend auf den NoSta-D-Annotation-Guidelines aus dem bereitgestellten Text zu extrahieren und ausschließlich als valides JSON-Objekt zurückzugeben.

  \#\#\# Entitätstypen:
  * **\texttt{<tag-name>}**: \texttt{<List of types for this tag also including some concrete examples>}
  * **location**: Geografische Orte, Länder, Staaten, Städte, Regionen, Straßen, Meere oder Planeten (z. B. "Stuttgart", "Hessen", "Frankreich").
  * ...
  
  \#\#\# Regeln:
  1. Extrahiere die Wörter exakt so, wie sie im Text stehen. Modifiziere oder korrigiere die Sätze oder Schreibweisen nicht.
  2. Wenn eine Entität im Text nicht vorkommt, gib ein leeres Array [] für diesen Schlüssel an.
  3. WICHTIG: Das JSON muss absolut valide sein. Setze NIEMALS ein Komma nach dem letzten Element ("person\_part"). Vermeide jegliche Syntaxfehler.

  \#\#\# Ziel-Format:
  ```json
  {
    \texttt{<tag-name>}: \texttt{[<annotated span>]},
    "location": ["Wert für location"],
    ...
  }
  ```

  \#\#\# Eingabetext:
  "{{input\_text}}"

  \#\#\# Extrahierte NER-Tags im JSON-Format:

\end{quote}

\section{Macro F1 Results}
\label{app:macro}

\begin{table*}[t]
  \centering
  \resizebox{\textwidth}{!}{%
  \begin{tabular}{l ccc ccc ccc c}
  \toprule
   & \multicolumn{3}{c}{\textbf{Ministral 8B}} & \multicolumn{3}{c}{\textbf{Qwen 3.5 9B}}
   & \multicolumn{3}{c}{\textbf{Gemma 4 12B}} & \textbf{GPT-5.4 mini} \\
  \cmidrule(lr){2-4} \cmidrule(lr){5-7} \cmidrule(lr){8-10} \cmidrule(lr){11-11}
  \textbf{Dataset} & 0-shot & 5-shot & LoRA & 0-shot & 5-shot & LoRA & 0-shot & 5-shot & LoRA & 0-shot \\
  \midrule 

FAIRagro & 0.517 & 0.456 & \underline{0.650} & 0.597 & 0.637 & \textbf{0.717} & 0.462 & 0.506 & 0.627 & 0.487 \\
Archaeo NER & 0.367 & 0.398 & 0.623 & 0.450 & 0.447 & \textbf{0.642} & 0.489 & 0.521 & \underline{0.636} & 0.466 \\
CO-Fun & 0.685 & 0.724 & \textbf{0.875} & 0.666 & 0.715 & \underline{0.853} & 0.661 & 0.739 & 0.836 & 0.649 \\
WikiBiography & 0.359 & 0.357 & \textbf{0.730} & 0.433 & 0.450 & 0.633 & 0.442 & 0.477 & \underline{0.679} & 0.457 \\
Europarl-ner & 0.356 & 0.403 & 0.719 & 0.412 & 0.496 & \underline{0.779} & 0.698 & 0.720 & \textbf{0.788} & 0.633 \\
SwissNER & 0.774 & -- & -- & \underline{0.806} & -- & -- & 0.790 & -- & -- & \textbf{0.860} \\
NoSta-D & 0.191 & 0.196 & 0.643 & 0.297 & 0.311 & \textbf{0.672} & 0.312 & 0.302 & \underline{0.669} & 0.316 \\
EHRI-NER & 0.551 & 0.596 & \textbf{0.828} & 0.584 & 0.587 & \underline{0.817} & 0.613 & 0.646 & 0.816 & 0.659 \\
NER Staatsarchiv & 0.595 & 0.643 & 0.849 & 0.699 & 0.715 & \textbf{0.883} & 0.737 & 0.784 & \underline{0.878} & 0.794 \\
German LER & 0.216 & 0.246 & \textbf{0.633} & 0.172 & 0.297 & \underline{0.481} & 0.241 & 0.180 & 0.328 & 0.274 \\
GerPS & 0.113 & 0.117 & 0.274 & 0.147 & 0.159 & \textbf{0.305} & 0.148 & 0.169 & \underline{0.284} & 0.135 \\
GERNERMED & 0.463 & 0.541 & 0.778 & 0.482 & 0.595 & \textbf{0.954} & 0.483 & 0.590 & \underline{0.875} & 0.504 \\
HUMADEX German NER & 0.213 & 0.247 & \textbf{0.664} & 0.240 & 0.239 & \underline{0.658} & 0.253 & 0.268 & 0.618 & 0.258 \\ \midrule
Average & 0.385 & 0.410 & \underline{0.689} & 0.431 & 0.471 & \textbf{0.700} & 0.462 & 0.492 & 0.669 & 0.469 \\

  \bottomrule
  \end{tabular}%
  }\vspace{-5px}
  \caption{Macro-averaged entity-level F1 on the German NER benchmark. Per dataset,
  the best score is \textbf{bold} and the second best \underline{underlined}.
  \emph{0-shot}: zero-shot, \emph{5-shot}: 5-shot in-context, \emph{LoRA}: fine-tuned.
  The final row reports per-column averages over the datasets, excluding SwissNER, which is only evaluated in the 0-shot setting due to the small dataset size that does not allow for creating train or validation splits.}
  \label{tab:ner-macro-f1}
  \end{table*}

\begin{table*}[t]
  \centering
  \resizebox{\textwidth}{!}{%
  \begin{tabular}{l ccc ccc ccc c}
  \toprule
   & \multicolumn{3}{c}{\textbf{Ministral 8B}} & \multicolumn{3}{c}{\textbf{Qwen 3.5 9B}}
   & \multicolumn{3}{c}{\textbf{Gemma 4 12B}} & \textbf{GPT-5.4 mini} \\
  \cmidrule(lr){2-4} \cmidrule(lr){5-7} \cmidrule(lr){8-10} \cmidrule(lr){11-11}
  \textbf{Dataset} & 0-shot & 5-shot & LoRA & 0-shot & 5-shot & LoRA & 0-shot & 5-shot & LoRA & 0-shot \\
  \midrule
FAIRagro & 0.723 & 0.629 & 0.723 & 0.737 & 0.687 & 0.717 & \underline{0.748} & 0.730 & \textbf{0.752} & 0.660 \\
Archaeo NER & 0.504 & 0.526 & \underline{0.711} & 0.560 & 0.560 & \textbf{0.723} & 0.588 & 0.625 & 0.705 & 0.578 \\
CO-Fun & 0.844 & 0.860 & \textbf{0.909} & 0.783 & 0.821 & \underline{0.891} & 0.805 & 0.868 & 0.871 & 0.774 \\
WikiBiography & 0.472 & 0.472 & \textbf{0.748} & 0.505 & 0.526 & 0.652 & 0.499 & 0.534 & \underline{0.698} & 0.515 \\
Europarl-ner & 0.526 & 0.545 & 0.799 & 0.616 & 0.628 & 0.800 & 0.792 & \underline{0.800} & \textbf{0.878} & 0.716 \\
SwissNER & 0.790 & -- & -- & \underline{0.820} & -- & -- & 0.804 & -- & -- & \textbf{0.862} \\
NoSta-D & 0.241 & 0.236 & 0.650 & 0.335 & 0.344 & \textbf{0.679} & 0.414 & 0.358 & \underline{0.678} & 0.354 \\
EHRI-NER & 0.738 & 0.777 & \textbf{0.883} & 0.720 & 0.714 & 0.852 & 0.740 & 0.772 & \underline{0.868} & 0.748 \\
NER Staatsarchiv & 0.801 & 0.818 & 0.886 & 0.848 & 0.835 & \textbf{0.915} & 0.807 & 0.875 & \underline{0.912} & 0.865 \\
German LER & 0.363 & 0.389 & \textbf{0.744} & 0.269 & 0.445 & \underline{0.618} & 0.390 & 0.233 & 0.412 & 0.467 \\
GerPS & 0.236 & 0.231 & 0.334 & 0.252 & 0.271 & \textbf{0.368} & 0.278 & 0.299 & \underline{0.341} & 0.273 \\
GERNERMED & 0.768 & 0.799 & 0.890 & 0.808 & 0.821 & \textbf{0.973} & 0.791 & 0.811 & \underline{0.939} & 0.823 \\
HUMADEX German NER & 0.486 & 0.508 & \textbf{0.782} & 0.500 & 0.501 & \underline{0.780} & 0.554 & 0.557 & 0.753 & 0.546 \\  \midrule
Average & 0.559 & 0.566 & \textbf{0.755} & 0.578 & 0.596 & \underline{0.747} & 0.617 & 0.622 & 0.734 & 0.610 \\
  \bottomrule
  \end{tabular}%
  }\vspace{-5px}
  \caption{Partial-match (fuzzy) entity-level macro F1 on the German NER benchmark, using
  token-set-ratio matching at threshold 0.90. Per dataset, best \textbf{bold},
  second best \underline{underlined}; final row = per-column averages.}
  \label{tab:ner-partial-macro-f1}
  \end{table*}\vspace{-5px}

\end{document}